\documentclass[letterpaper]{article}
\usepackage{aaai}
\usepackage{times}
\usepackage{helvet}
\usepackage{courier}
\usepackage{graphicx}
\usepackage{balance}
\usepackage{tikz}
\usepackage{amsmath}
\usepackage[justification=centering]{caption}
\usetikzlibrary{shapes.geometric, arrows}
\tikzstyle{io} = [rectangle, rounded corners, text centered, text height=110, text width=75, draw=black, fill=blue!15]
\tikzstyle{layer} = [rectangle, rounded corners, text centered, text height=4, text width=70, draw=black, fill=yellow!40]
\tikzstyle{avg} = [rectangle, rounded corners, text centered, text height=4, text width=20, draw=black, fill=white]
\newcommand{\src}{\scriptsize init. from }
\begin{document}
%
\title{CAMeMBERT: Cascading Assistant-Mediated Multilingual BERT}
\author{Dan DeGenaro\\
University of Massachusetts, Amherst\\
Department of Linguistics\\
650 North Pleasant St.\\
Amherst, MA 01003\\
ddegenaro@umass.edu\\
\And Jugal Kalita\\
University of Colorado, Colorado Springs\\
Department of Computer Science\\
1420 Austin Bluffs Pkwy\\
Colorado Springs, CO 80918\\
jkalita@uccs.edu\\
}
\maketitle
\begin{abstract}
\begin{quote}
Large language models having hundreds of millions, and even billions, of parameters have performed extremely well on a variety of natural language processing (NLP) tasks. Their widespread use and adoption, however, is hindered by the lack of availability and portability of sufficiently large computational resources. This paper proposes a knowledge distillation (KD) technique building on the work of LightMBERT, a student model of multilingual BERT (mBERT). By repeatedly distilling mBERT through increasingly compressed top-layer distilled teacher assistant networks, CAMeMBERT aims to improve upon the time and space complexities of mBERT while keeping loss of accuracy beneath an acceptable threshold. At present, CAMeMBERT has an average accuracy of around 60.1\%, which is subject to change after future improvements to the hyperparameters used in fine-tuning.
\end{quote}
\end{abstract}

\section{Introduction}
\noindent Large multilingual language models such as multilingual BERT (mBERT) have excelled at tasks such as machine translation, question answering, and structured predictions \cite{hu_xtreme_2020}. However, they are generally too computationally expensive to be used on personal devices. A solution to this problem is knowledge distillation (KD). KD was first suggested by \citeauthor{bucilua_model_2006} \shortcite{bucilua_model_2006}, but was first popularized among machine learning researchers by \citeauthor{hinton_distilling_2015} \shortcite{hinton_distilling_2015}. KD refers to the idea of ``distilling" a larger model into a much smaller one; the models are often called the ``teacher" and ``student" networks, respectively. Many successful KD techniques have been employed in order to reduce the computational needs of these large neural networks, while maintaining relatively small losses in accuracy. A simple approach to KD may be using a loss function that minimizes the difference between the logits of the teacher and student models.
\begin{itemize}
    \item This paper aims to improve upon the accuracy of \cite{jiao_lightmbert_2021} by applying a similar distillation technique and additionally using teacher assistant networks as described by \cite{mirzadeh_improved_2020}.
    \item In so doing, this paper also proposes use of adjacent layer averaging as a teacher-to-student layer mapping during the distillation process.
\end{itemize}
What follows is a brief review of related work (Section 2), a fuller description of both the problem at hand (Section 3) and our approach to solving it (Section 4), and an evaluation of our model's performance on the XNLI (Cross-lingual Natural Language Inference) metric \cite{conneau_xnli_2018}, a common benchmark for multilingual language models (Section 5).

\section{Related Work}
\subsection{Initializing from Teacher Networks}
Rather than directly training the student model on the training data using tasks like masked language modeling, it has been shown to be more effective to train the student model to just mimic the logits, hidden layers, or attention matrices of the teacher \cite{gou_knowledge_2021}. This has been done in several ways. One of the first methods employed, DistilBERT \cite{sanh_distilbert_2020}, involved training a new, smaller network with half as many layers to mimic the original model. DistilBERT performed extremely well on various benchmarks (such as CoLA, MNLI, and SQuAD), and was initialized with every other layer of its teacher. This approach led to a slightly different student initialization method known as ``top-layer distillation" \cite{jiao_lightmbert_2021}, in which the student network is initialized with the lower layers of the teacher network, and trained in similar fashion to mimic the teacher. In \citeauthor{jiao_lightmbert_2021}'s paper, the student network, LightMBERT, was initialized with the lower six encoder layers of mBERT, and then trained to mimic the teacher mBERT.
\subsection{Teacher Assistant Networks}
In computer vision, an increasingly popular framework for KD makes use of ``teacher assistant" (TA) networks that try to bridge the large gap in capabilities between student and teacher networks \cite{mirzadeh_improved_2020}. Teacher Assistant Knowledge Distillation (TAKD) has been shown to improve retention of information by student networks. Indeed, Mirzadeh et al. show that a distillation path having the maximal number of TAs (removing only one or two layers at a time) produces optimal results.

\begin{figure*}
\centering
\begin{tikzpicture}[node distance=1cm]
\node (src) at (0,0) [io] {Teacher};
\node (src7) at (0,1.7) [layer] {$ {\scriptstyle h_m} $};
\node (src6) at (0,1.2) [layer] {$ {\scriptstyle \hdots} $};
\node (src5) at (0,0.7) [layer] {$ {\scriptstyle h_5} $};
\node (src4) at (0,0.2) [layer] {$ {\scriptstyle h_4} $};
\node (src3) at (0,-0.3) [layer] {$ {\scriptstyle h_3} $};
\node (src2) at (0,-0.8) [layer] {$ {\scriptstyle h_2} $};
\node (src1) at (0,-1.3) [layer] {$ {\scriptstyle h_1} $};

\node (ta1) at (3,0) [io] {TA$_1$};
\node (ta16) at (3,1.2) [layer] {\src Teacher};
\node (ta15) at (3,0.7) [layer] {$ {\scriptstyle \hdots} $};
\node (ta14) at (3,0.2) [layer] {\src Teacher};
\node (ta13) at (3,-0.3) [layer] {\src Teacher};
\node (ta12) at (3,-0.8) [layer] {\src Teacher};
\node (ta11) at (3,-1.3) [layer] {\src Teacher};

\node (ta2) at (6,0) [io] {TA$_2$};
\node (ta25) at (6,0.7) [layer] {\src TA$_1$};
\node (ta24) at (6,0.2) [layer] {$ {\scriptstyle \hdots} $};
\node (ta23) at (6,-0.3) [layer] {\src TA$_1$};
\node (ta22) at (6,-0.8) [layer] {\src TA$_1$};
\node (ta21) at (6,-1.3) [layer] {\src TA$_1$};

\node (dots4) at (9,0.2) {$\hdots$};
\node (dots3) at (9,-0.3) {$\hdots$};
\node (dots2) at (9,-0.8) {$\hdots$};
\node (dots1) at (9,-1.3) {$\hdots$};

\node (tan) at (12,0) [io] {TA$_n$};
\node (tan3) at (12,-0.3) [layer] {\src TA$_{n-1}$};
\node (tan2) at (12,-0.8) [layer] {\src TA$_{n-1}$};
\node (tan1) at (12,-1.3) [layer] {\src TA$_{n-1}$};

\node (c) at (15,0) [io] {Student};
\node (c2) at (15,-0.8) [layer] {\src TA$_{n}$};
\node (c1) at (15,-1.3) [layer] {\src TA$_{n}$};

\draw[->, ultra thick] (src1) to (ta11);
\draw[->, ultra thick] (src2) to (ta12);
\draw[->, ultra thick] (src3) to (ta13);
\draw[->, ultra thick] (src4) to (ta14);
\draw[->, ultra thick] (src5) to (ta15);
\draw[->, ultra thick] (src6) to (ta16);

\draw[->, ultra thick] (ta11) to (ta21);
\draw[->, ultra thick] (ta12) to (ta22);
\draw[->, ultra thick] (ta13) to (ta23);
\draw[->, ultra thick] (ta14) to (ta24);
\draw[->, ultra thick] (ta15) to (ta25);

\draw[->, ultra thick] (ta21) to (dots1);
\draw[->, ultra thick] (ta22) to (dots2);
\draw[->, ultra thick] (ta23) to (dots3);
\draw[->, ultra thick] (ta24) to (dots4);

\draw[->, ultra thick] (dots1) to (tan1);
\draw[->, ultra thick] (dots2) to (tan2);
\draw[->, ultra thick] (dots3) to (tan3);

\draw[->, ultra thick] (tan1) to (c1);
\draw[->, ultra thick] (tan2) to (c2);

\end{tikzpicture}
\caption{Repeated top-layer distillation via TA networks (embeddings not shown). The number of hidden layers in network $j$ is one fewer than in network $j-1$.}
\end{figure*}
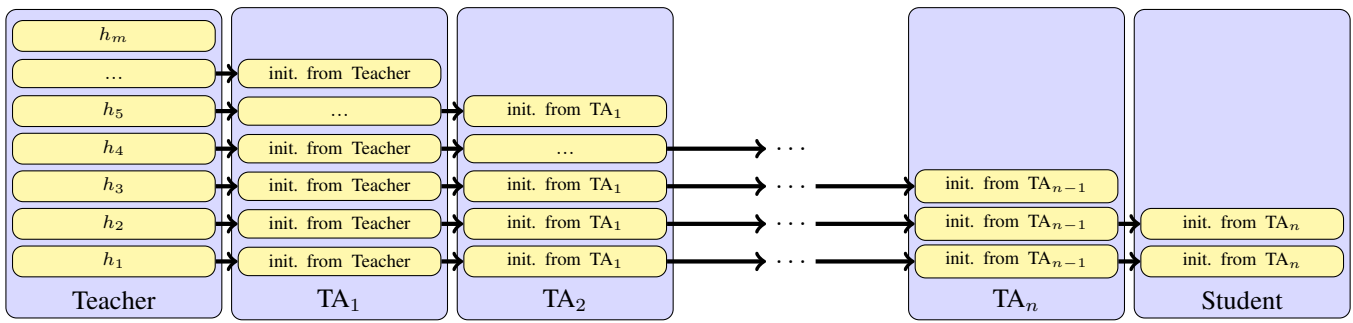

\section{Approach}
Our approach combines several techniques that have been successfully employed independently, namely top-layer distillation \cite{jiao_lightmbert_2021} and iterated TA networks \cite{mirzadeh_improved_2020}. CAMeMBERT was constructed via iterated top-layer distilled TA networks. Starting from the 12-layer mBERT network, a TA network is initialized with the lowest 11 layers of mBERT, including both weights and layer architecture (cutting out the topmost layer). This TA network is then trained to mimic the teacher's (mBERT's) hidden outputs and attention matrices, using mean-squared error (MSE) loss. Then, a second TA network is constructed using the first, again with the top encoder layer removed. The second network is initialized with the lowest ten layers of the first TA, both weights and layer architecture. This process is iterated until a network of six hidden layers is left, which is the CAMeMBERT model. This model is the same size as LightMBERT \cite{jiao_lightmbert_2021}. See Figure 1 for a diagram of this process.

\subsection{Training Details}
\subsubsection{Pretraining Data}
The training corpus consists of lines of text scraped from the largest 104 languages on Wikipedia\footnote{The text was obtained via HuggingFace's Datasets library, except for the languages Cebuano and Spanish, which were obtained via the WikiExtractor tool (all from the 2022-03-01 dump).} \cite{wikidump}, as described by \citeauthor{devlin_bert_2019} \shortcite{devlin_bert_2019}. The Wikipedias were concatenated into a single file, though parts of the file were not used.

The languages were sampled by a probability distribution $P'$ defined as

\begin{equation}
P'(\text{lang}_j) = \frac{P(\text{lang}_j)^S}{\sum_kP(\text{lang}_k)^S}
\end{equation}

\noindent
where $P(\text{lang}_j)$ denotes the probability of language $j$ according to the proportional space it occupies on disk, i.e.

\begin{equation}
P(\text{lang}_j) = \frac{\text{size}(\text{lang}_j)}{\sum_k\text{size}(\text{lang}_k)}
\end{equation}

\noindent
$S$ was chosen such that $P'(\text{English}) = 100P'(\text{Icelandic})$, as described by the BERT/mBERT team\footnote{Consider looking at this markdown file in the GitHub repo for BERT, which gives more details: https://github.com/google-research/bert/blob/master/multilingual.md}. The texts used to produce the corpus were cleaned to contain only the plain text of an article, including headings, and blank lines were removed. The articles from all languages were then concatenated into one large text file, whose lines were rearranged in random order using the GNU/Linux \texttt{shuf} utility. Each line was considered one training example (so a batch size of 256 would involve accumulating loss over 256 lines, one at a time, before performing an optimizer step).

\begin{table}[h!]
    \caption{Language abbreviations and representation.}
    \centering
    \begin{tabular}{c|c|c}
    Language & Abbreviation & \% of corpus \\
    \hline\hline
    English  & en           & 12.0 \\
    \hline
    Spanish  & es           & 3.49 \\
    \hline
    Chinese  & zh           & 2.08 \\
    \hline
    German   & de           & 6.06 \\
    \hline
    Arabic   & ar           & 2.33 \\
    \hline
    Urdu     & ur           & 0.365 \\
    \end{tabular}
    \label{tab:my_label}
\end{table}

\subsubsection{Pretraining Method}
Each network was trained for 66,666 steps (batches), for a total of about 400,000 steps across all six networks. Each pretraining distillation process used a distinct section of the corpus, unseen by preceding or following networks. The mBERT pretrained tokenizer from HuggingFace was used. This tokenizer is cased, and at no point was lowercasing was performed on the text. The vocabulary size for this tokenizer is 119,547. Maximum length padding was employed, along with truncation, with a maximum input sequence length of 128. All networks were kept in training mode (i.e. dropout enabled) throughout the process, and embedding layers were always frozen. Each network was optimized via an Adam optimizer defined by the following hyperparameters: batch size 256, peak learning rate $1\times 10^{-7}$, linear warmup over the first 6,666 steps, linear decay, $\beta_{1,2} = 0.9,0.999$, no weight decay, and $\epsilon = 1\times 10^{-9}$. Linear warmup was applied throughout the training process for the 11-layer network, because loss decreased more reliably under this setting. These choices are summarized in Table 2.

The total loss for a batch is defined as follows, following \citeauthor{jiao_lightmbert_2021} \shortcite{jiao_lightmbert_2021}:

\begin{equation}
\mathcal{L} = \frac{1}{n}\bigg(\sum_{j=1}^n \mathcal{L}^A_j + \sum_{k=1}^{n+1}\mathcal{L}^H_k\bigg)
\end{equation}

\noindent
where the student network has $n$ attention layers and $n+1$ hidden outputs (this implies the teacher has $n+1$ attention layers and $n+2$ hidden outputs). $\mathcal{L}^H_k$ is the loss of one layer's hidden outputs, and $\mathcal{L}^A_j$ is the loss of one layer's attentions. These are defined thus:

\begin{equation}
\mathcal{L}^A_j = \frac{1}{12}\sum_{\ell=1}^{12}\text{MSE}\bigg( \frac{1}{2}\bigg(A_T^{j\ell} + A_T^{j+1,\ell}\bigg) , A_S^{j\ell} \bigg)
\end{equation}

\begin{equation}
\mathcal{L}^H_k = \text{MSE}\bigg( \frac{1}{2}\bigg(H_T^k + H_T^{k+1}\bigg) , H_S^k \bigg)
\end{equation}

\noindent
where $_T,_S$ denote teacher and student, respectively, and $\ell$ ranges over the attention heads, of which there are 12 in the case of a BERT$_{\text{BASE}}$-style network. $\mathcal{L}_j^A$ is therefore the average loss of attention over the 12 attention heads at the $j$-th layer.

It should be noted that $\mathcal{L}_j^A$ and $\mathcal{L}_k^H$ are defined differently from \citeauthor{jiao_lightmbert_2021} \shortcite{jiao_lightmbert_2021} out of necessity; in that paper, a ``layer mapping" can be performed from 12 layers to six layers directly, but in this paper, since one layer is removed at a time, the mapping must map 12 to 11, 11 to ten, and so on. Thus, the mapping is defined (as is evident from the definition of the loss at one layer) to be the average of layers $j$ and $j+1$ of the teacher to the $j$-th layer of the student, a strategy we call ``adjacent layer averaging." To be clear, the average of layers six and seven of a teacher would be used to train the sixth layer of its student, for instance. See Figure 2 for a diagram of this process.

\begin{figure*}
\centering
\begin{tikzpicture}[node distance=1cm]
\node (src) at (0,0) [io] {TA$_5$ (7 layers)};
\node (src7) at (0,1.7) [layer] {$ {\scriptstyle h_7} $};
\node (src6) at (0,1.2) [layer] {$ {\scriptstyle h_6} $};
\node (src5) at (0,0.7) [layer] {$ {\scriptstyle h_5} $};
\node (src4) at (0,0.2) [layer] {$ {\scriptstyle h_4} $};
\node (src3) at (0,-0.3) [layer] {$ {\scriptstyle h_3} $};
\node (src2) at (0,-0.8) [layer] {$ {\scriptstyle h_2} $};
\node (src1) at (0,-1.3) [layer] {$ {\scriptstyle h_1} $};

\node (avg6) at (3,1.2) [avg] {AVG};
\node (avg5) at (3,0.7) [avg] {AVG};
\node (avg4) at (3,0.2) [avg] {AVG};
\node (avg3) at (3,-0.3) [avg] {AVG};
\node (avg2) at (3,-0.8) [avg] {AVG};
\node (avg1) at (3,-1.3) [avg] {AVG};

\node (ta1) at (6,0) [io] {CAMeMBERT};
\node (ta16) at (6,1.2) [layer] {$ {\scriptstyle h_6} $};
\node (ta15) at (6,0.7) [layer] {$ {\scriptstyle h_5} $};
\node (ta14) at (6,0.2) [layer] {$ {\scriptstyle h_4} $};
\node (ta13) at (6,-0.3) [layer] {$ {\scriptstyle h_3} $};
\node (ta12) at (6,-0.8) [layer] {$ {\scriptstyle h_2} $};
\node (ta11) at (6,-1.3) [layer] {$ {\scriptstyle h_1} $};

\draw[->, ultra thick] (src1) to (avg1);
\draw[->, ultra thick] (src2) to (avg2);
\draw[->, ultra thick] (src3) to (avg3);
\draw[->, ultra thick] (src4) to (avg4);
\draw[->, ultra thick] (src5) to (avg5);
\draw[->, ultra thick] (src6) to (avg6);

\draw[->, ultra thick] (src2) to (avg1);
\draw[->, ultra thick] (src3) to (avg2);
\draw[->, ultra thick] (src4) to (avg3);
\draw[->, ultra thick] (src5) to (avg4);
\draw[->, ultra thick] (src6) to (avg5);
\draw[->, ultra thick] (src7) to (avg6);

\draw[->, ultra thick] (avg1) to (ta11);
\draw[->, ultra thick] (avg2) to (ta12);
\draw[->, ultra thick] (avg3) to (ta13);
\draw[->, ultra thick] (avg4) to (ta14);
\draw[->, ultra thick] (avg5) to (ta15);
\draw[->, ultra thick] (avg6) to (ta16);

\end{tikzpicture}
\caption{Adjacent layer averaging between a teacher network and its student. The number of hidden layers in network $j$ is one fewer than in network $j-1$.}
\end{figure*}
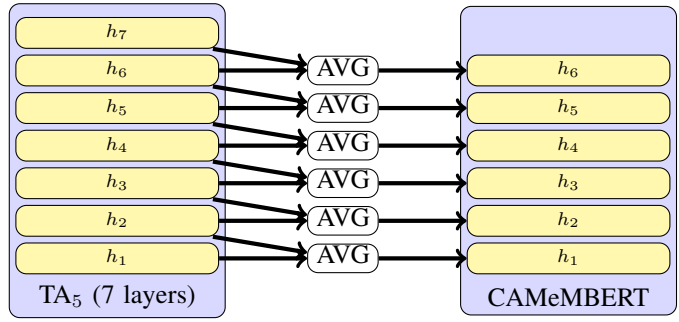

It is also important to define MSE clearly. MSE refers to mean squared error, which is defined generically as:

\begin{equation}
\text{MSE}(X, Y) = \frac{1}{\text{numel}(X)}\sum_{j,k,...}(X_{jk...} - Y_{jk...})^2
\end{equation}

\noindent
where $\text{shape}(X) = \text{shape}(Y)$, both can be indexed by indices $j, k, ...$ and $\text{numel}(X)$ (which equals $\text{numel}(Y)$) denotes the number of elements in $X$ (product of ranges of all indices).

\subsubsection{Fine-tuning Method}
The networks were fine-tuned on the English XNLI dataset's ``train" split obtained via HuggingFace's Datasets library, following \citeauthor{jiao_lightmbert_2021} \shortcite{jiao_lightmbert_2021}. Fine-tuning was conducting by an Adam optimizer as follows: 3 epochs, batch size 32, learning rate $2\times 10^{-5}$, $\beta_{1,2} = 0.9,0.999$, no weight decay, and $\epsilon = 2\times 10^{-7}$. Padding and truncation were applied with a maximum sequence length of 128. Our network's weights were loaded into a HuggingFace BertForSequenceClassification architecture, with weights of the fine-tuning layers being randomly initialized. The number of output classes was set to be 3, as XNLI data are labeled as either `entailment' (0), `neutral' (1), or `contradiction' (2). Cross entropy loss was employed as the fine-tuning loss\footnote{See documentation here: https://pytorch.org/docs/stable/\\generated/torch.nn.CrossEntropyLoss.html}.

\begin{table}[h!]
\centering
\caption{Hyperparameters. Note that the abbreviation `LR' refers to learning rate.}
\begin{tabular}{c|c|c}
Hyperparameter & Pretraining & Fine-tuning \\
\hline\hline
optimizer & Adam & Adam \\
\hline
epochs & 1 & 3 \\
\hline
steps & 66,666 & 12,271 \\
\hline
batch size & 256 & 32 \\
\hline
peak LR & $1\times 10^{-7}$ & $2\times 10^{-5}$ \\
\hline
LR warmup & linear, first 6,666 steps* & none \\
\hline
LR decay & linear, to the end & none \\
\hline
$\beta_1$ & 0.9 & 0.9 \\
\hline
$\beta_2$ & 0.999 & 0.999 \\
\hline
weight decay & 0 & 0 \\
\hline
$\epsilon$ & $1\times 10^{-9}$ & $2\times 10^{-7}$ \\
\hline
padding & True & True \\
\hline
max seq. length & 128 & 128 \\
\hline
embeddings frozen & True & True \\
\hline
vocab size & 119,547 & 119,547 \\
\end{tabular}
\caption*{*The first TA (11 layer network) had linear warmup over all 66,666 steps.}
\end{table}

\subsection{Evaluation Metric}
The networks were evaluated on the zero-shot cross-lingual transfer task using HuggingFace's ``test" split of the XNLI dataset. It was evaluated on the six languages present in Table 1 for comparison with the results provided by the LightMBERT team \cite{jiao_lightmbert_2021}, and the mean accuracy (AVG) displayed in Table 3 is the mean accuracy across those six languages.

\section{Results}
At present, the networks do not perform as well as LightMBERT, but close to it (60.1\% vs. 70.3\% on average across the six languages shown in Table 3). Extensive experiments were carried out in terms of tuning both pretraining and fine-tuning hyperparameters, and the results presented are the best outcomes that this technique was able to produce. It is unclear why there is such a large drop in accuracy each time a layer is removed, almost on par with not conducting further pretraining at all. The results may be poor in part due to the mixture of training data languages, which is subject to change significantly as people contribute articles to Wikipedia in various languages.

\balance

\section{Conclusion}
By applying repeated TA-mediated top-layer distillations to a large language model, this work stands to produce a fast, memory- and storage-efficient neural network that can mimic the abilities of mBERT. This work builds on that of LightMBERT, which was created via top-layer distillation, as well as that of the TAKD framework for more effective knowledge distillation. Although this work was unable to improve upon these techniques, this may be due to a practical issue, rather than a theoretical issue. We believe that the reasoning of this project was sound, and as such, we encourage future research into this technique and other techniques in similar veins.

\section{Acknowledgement}
The work reported in this paper is supported by the National Science Foundation under Grant No. 2050919. Any opinions, findings and conclusions or recommendations expressed in this work are those of the author(s) and do not necessarily reflect the views of the National Science Foundation. Thanks to Abigail Swenor (Notre Dame), Aaron Serianni (Princeton), and Dr. Terrance Boult (UCCS) for their help with the practical side of this project.

\begin{center}
\begin{table}[h!]
\caption{Results on XNLI Task by Language and Number of Layers}
\begin{tabular}{c|c|c|c|c|c|c|c}
   &  en  &  es  &  zh  &  de  &  ar  &  ur  & AVG \\
\hline
11 & 81.0 & 73.3 & 69.3 & 70.1 & 64.6 & 57.1 & 69.2 \\
\hline
10 & 79.2 & 71.3 & 66.1 & 67.0 & 61.8 & 56.4 & 67.0 \\
\hline
9  & 79.9 & 69.6 & 65.9 & 65.5 & 60.9 & 56.4 & 66.4 \\
\hline
8  & 77.8 & 67.8 & 64.9 & 64.2 & 58.9 & 54.5 & 64.7 \\
\hline
7  & 77.7 & 65.7 & 64.6 & 62.3 & 57.1 & 52.3 & 63.3 \\
\hline
6  & 76.8 & 62.1 & 60.7 & 60.1 & 51.5 & 49.6 & 60.1 \\
\hline\hline
LMB* & 81.5 & 74.7 & 69.3 & 72.2 & 65.0 & 59.3 & 70.3 \\
\end{tabular}
\caption*{*LMB is LightMBERT's results \cite{jiao_lightmbert_2021}. Abbreviations explained in Table 1.}
\end{table}
\end{center}

\bibliographystyle{aaai}
\bibliography{aaai}
\end{document}